\documentclass[fleqn,10pt]{wlscirep}

\usepackage{tcolorbox}
\usepackage{amsmath}
\usepackage{enumitem} %

\usepackage[utf8]{inputenc}
\usepackage{lipsum}
\usepackage[T1]{fontenc}
\usepackage{bm}
\usepackage{pgfplots} %
\usepackage{float} %
\usepackage{etoolbox}
\usepackage{datetime2}
\usepackage{todonotes}

\usepackage[super,sort&compress]{natbib}
\setcitestyle{citesep={,}}
\usepackage{bibunits}
\defaultbibliographystyle{naturemag-doi}
\defaultbibliography{references}

\definecolor{myblue}{rgb}{0.0, 0.0, 0.99}
\definecolor{myblue}{rgb}{0, 0, 0}
\definecolor{red}{rgb}{0, 0, 0}

\usepackage{shellesc} %

\newif\ifcountwords
\countwordstrue
\countwordsfalse

\newcommand{\wordcounttext}[1]{%
  \ifcountwords
        \leavevmode\textbf{{\color{blue} Word count for #1}}%
  \fi
}

\usepackage{multirow}
\usepackage{makecell}
\usepackage{longtable}

\usepackage{graphicx}
\usepackage[normalem]{ulem}
\usepackage{lscape}
\usepackage{pdflscape}
\usepackage{afterpage}

\usepackage{lineno}

\def\SectionName{sections-initial-draft}
\def\SectionName{SECTIONS-FOR-SUBMISSION}

\title{Towards deployment-centric multimodal AI beyond vision and language}%

\author[1,32]{Xianyuan Liu}
\author[1,32]{Jiayang Zhang}
\author[2,32]{Shuo Zhou}
\author[3]{Thijs L. van der Plas}
\author[4]{Avish Vijayaraghavan}
\author[5]{Anastasiia Grishina}
\author[6]{Mengdie Zhuang}
\author[7]{Daniel Schofield}
\author[8]{Christopher Tomlinson}
\author[9]{Yuhan Wang}
\author[10]{Ruizhe Li}
\author[3]{Louisa van Zeeland}
\author[2]{Sina Tabakhi}
\author[11]{Cyndie Demeocq}
\author[12]{Xiang Li}
\author[13]{Arunav Das}
\author[14]{Orlando Timmerman}
\author[15]{Thomas Baldwin-McDonald}
\author[8]{Jinge Wu}
\author[2]{Peizhen Bai}
\author[16]{Zahraa Al Sahili}
\author[17]{Omnia Alwazzan}
\author[18]{Thao N. Do}
\author[1]{Mohammod N.I. Suvon}
\author[19]{Angeline Wang}
\author[20]{Lucia Cipolina-Kun}
\author[21]{Luigi A. Moretti}
\author[22]{Lucas Farndale}
\author[13]{Nitisha Jain}
\author[23]{Natalia Efremova}
\author[12]{Yan Ge}
\author[24]{Marta Varela}
\author[9]{Hak-Keung Lam}
\author[9]{Oya Celiktutan}
\author[25]{Ben R. Evans}
\author[3]{Alejandro Coca-Castro}
\author[26]{Honghan Wu}
\author[12]{Zahraa S. Abdallah}
\author[2]{Chen Chen}
\author[23]{Valentin Danchev}
\author[27]{Nataliya Tkachenko}
\author[28]{Lei Lu}
\author[29]{Tingting Zhu}
\author[17]{Gregory G. Slabaugh}
\author[2]{Roger K. Moore}
\author[30]{William K. Cheung}
\author[31]{Peter H. Charlton}
\author[2,32,*]{Haiping Lu}

\affil[1]{Centre for Machine Intelligence, University of Sheffield, Sheffield, UK}
\affil[2]{School of Computer Science, University of Sheffield, Sheffield, UK}
\affil[3]{The Alan Turing Institute, London, UK}
\affil[4]{Department of Metabolism, Digestion and Reproduction, Imperial College London, London, UK}
\affil[5]{Department of Applied AI, Simula Research Laboratory, Oslo, Norway}
\affil[6]{Information School, University of Sheffield, Sheffield, UK}
\affil[7]{NHS England, Leeds, UK}
\affil[8]{Institute of Health Informatics, University College London, London, UK}
\affil[9]{Department of Engineering, King's College London, London, UK}
\affil[10]{Department of Computing Science, University of Aberdeen, Aberdeen, UK}
\affil[11]{School of Informatics, University of Edinburgh, Edinburgh, UK}
\affil[12]{School of Engineering Mathematics and Technology, University of Bristol, Bristol, UK}
\affil[13]{Department of Informatics, King's College London, London, UK}
\affil[14]{Department of Earth Sciences, University of Cambridge, Cambridge, UK}
\affil[15]{Department of Computer Science,  University of Manchester, Manchester, UK}
\affil[16]{Department of Computer Science, Queen Mary University of London, London, UK}
\affil[17]{Digital Environment Research Institute, Queen Mary University of London, London, UK }
\affil[18]{Department of Computer Science, University of Bath, Bath, UK}
\affil[19]{Department of Classics, King's College London, London, UK}
\affil[20]{School of Electrical, Electronic and Mechanical Engineering, University of Bristol, Bristol, UK}
\affil[21]{School of Engineering, University of the West of England, Bristol, UK}
\affil[22]{Cancer Research UK Scotland Institute, Glasgow, UK}
\affil[23]{School of Business and Management, Queen Mary University of London, London, UK}
\affil[24]{City St George's University of London, London, UK}
\affil[25]{British Antarctic Survey, Cambridge, UK}
\affil[26]{School of Health and Wellbeing, University of Glasgow, Glasgow, UK}
\affil[27]{Chief Data \& AI Office, Lloyds Banking Group, London, UK}
\affil[28]{School of Life Course \& Population Sciences, King's College London, London, UK}
\affil[29]{Institute of Biomedical Engineering, University of Oxford, Oxford, UK}
\affil[30]{Department of Computer Science, Hong Kong Baptist University, Hong Kong, China}
\affil[31]{Department of Public Health and Primary Care, University of Cambridge, Cambridge, UK}

\affil[32]{These authors contributed equally: Xianyuan Liu, Jiayang Zhang, Shuo Zhou, and Haiping Lu.}

\affil[*]{Corresponding author: Haiping Lu (h.lu@sheffield.ac.uk)}

\newpage 
\begin{abstract}
Multimodal artificial intelligence (AI) integrates diverse types of data via machine learning to improve understanding, prediction, and decision-making across disciplines such as healthcare, science, and engineering. However, most multimodal AI advances focus on models for vision and language data, while their deployability remains a key challenge. We advocate a deployment-centric workflow that incorporates deployment constraints early to reduce the likelihood of undeployable solutions, complementing data-centric and model-centric approaches. We also emphasise deeper integration across multiple levels of multimodality {\color{myblue}through stakeholder engagement and interdisciplinary collaboration} to broaden the research scope beyond vision and language. To facilitate this approach, we identify common multimodal-AI-specific challenges shared across disciplines and examine three real-world use cases: pandemic response, self-driving car design, and climate change adaptation, drawing expertise from healthcare, social science, engineering, science, sustainability, and finance. By fostering {\color{myblue}interdisciplinary} dialogue and open research practices, our community can accelerate deployment-centric development for broad societal impact.

\end{abstract}

\pgfplotsset{compat=1.18}
\begin{document}
\SetLipsumDefault{1}

\pagestyle{mainstyle}  %

\tcbset{
    naturebox/.style={
        colback=gray!10,             %
        colframe=black,              %
        boxrule=0.5pt,               %
        fonttitle=\bfseries,         %
        title={\large Box 1 \hspace{0.5em} Strategic recommendations for advancing multimodal AI across disciplines}, %
        colbacktitle=gray!10,        %
        coltitle=black,              %
        toptitle=1mm,                %
        bottomtitle=1mm,             %
        left=1mm,                    %
        right=1mm,                   %
        fontupper=\small             %
    }
}

\flushbottom
\maketitle

\wordcounttext{abstract:}

\begin{bibunit}

\begin{figure*}[!b]
\centering
    \includegraphics[width=\textwidth]{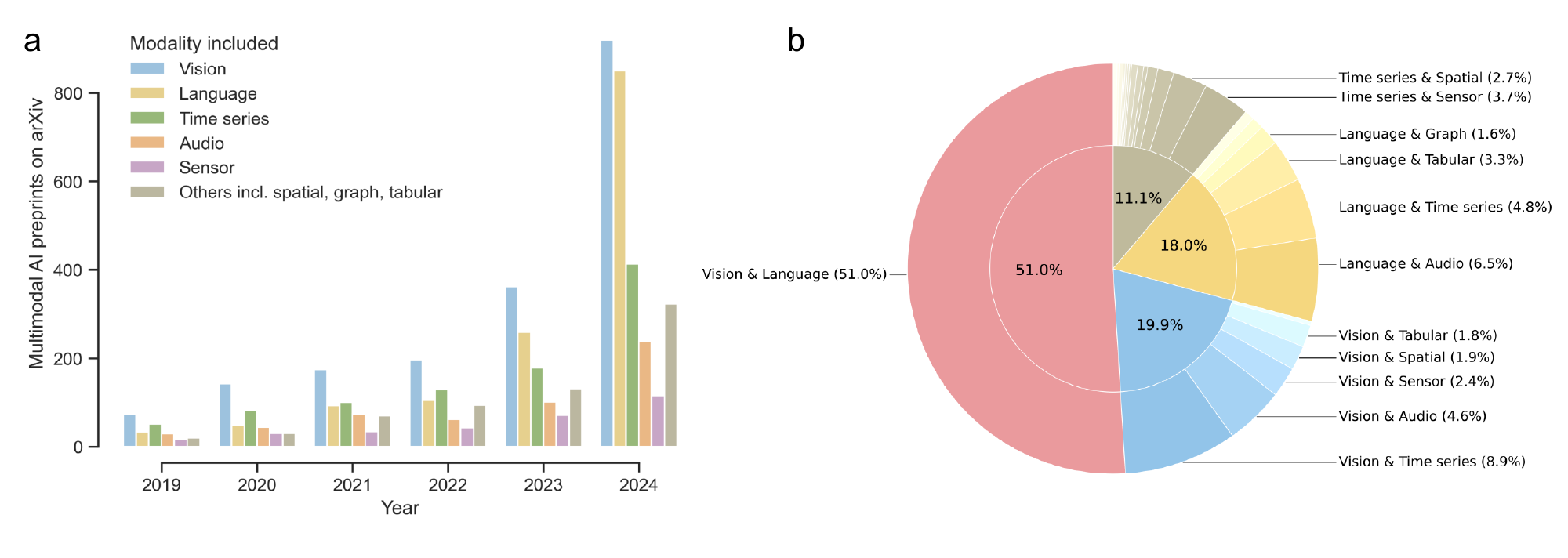}
    \caption{{\color{myblue}\textbf{Trends in multimodal AI research (2019--2024) and the dominance of vision and language.}} 
    \textbf{a,} Yearly growth of multimodal AI preprints on arXiv by modality, showing a steady increase over time and the dominance of vision and language. 
    \textbf{b,} Breakdown of modality pairs in multimodal AI preprints on arXiv in 2024, revealing that over half the studies involve vision and language, followed by vision and others (19.9\%), language and others (18.0\%), and other modality combinations (11.1\%). This analysis highlights the most common pairwise modality combinations and shows that those involving vision or language dominate, reaching 88.9\%. For clarity and space efficiency, only modality pairs exceeding 1.5\% are annotated. 
    {\color{myblue}See Supplementary Section S2 for methodology and details of the trend analysis. 
    See Supplementary Fig. \ref{fig:modality_combinations} for a more detailed illustration and analysis of the multimodal AI landscape, including statistics on triple and quadruple modality combinations. }}
\label{fig:mmai-trend}
\end{figure*}

\section*{Introduction}
Data drives discovery in the 21st century \cite{wang2023scientific}. From satellites monitoring climate change to social media capturing human behaviour, the variety and volume of information have never been greater. Each type of data, or \textit{modality}, offers unique insights, but unimodal approaches often fall short of achieving robust or generalisable performance. Autonomous vehicles relying solely on visual data struggle with object detection in low-light or adverse weather conditions \cite{kabir2025terrain}. During the COVID-19 pandemic, relying solely on RT-PCR data for diagnosing SARS-CoV-2 infection led to frequent false negatives, hindering timely interventions \cite{woloshin2020false}. Instead, integrating information from multiple modalities can reveal patterns and solutions that unimodal approaches miss \cite{Zhao2024DeepMD}.

Multimodal artificial intelligence (AI) leverages multimodal data for better understanding complex systems through machine learning  \cite{baltruvsaitis2018multimodal,ektefaie2023multimodal,xu2023multimodal,liang2024foundations}. Its promise is evident in multidisciplinary applications such as pandemic response. For example, in healthcare, combining medical imaging, genomic sequencing, and epidemiological data can enhance diagnoses, inform treatment strategies, and support disease prevention \cite{acosta2022multimodal, kline2022multimodal,krones2025review}. In science, fusing genetic and protein structure data holds promise for advancing vaccine development \cite{notin2024machine}. In engineering, integrating textual specifications with spatial data can improve product design and manufacturing \cite{song2024multi}, which could extend to optimising ventilator production. Across other disciplines, such as sustainability, finance, and social science, multimodal AI offers the potential to provide deeper insights and actionable strategies \cite{ofodile2024predictive, quatrini2021challenges, gupta2021emotion, anshul2023multimodal}.

\begin{figure*}[!t]
\centering
    \includegraphics[width=0.8\textwidth]{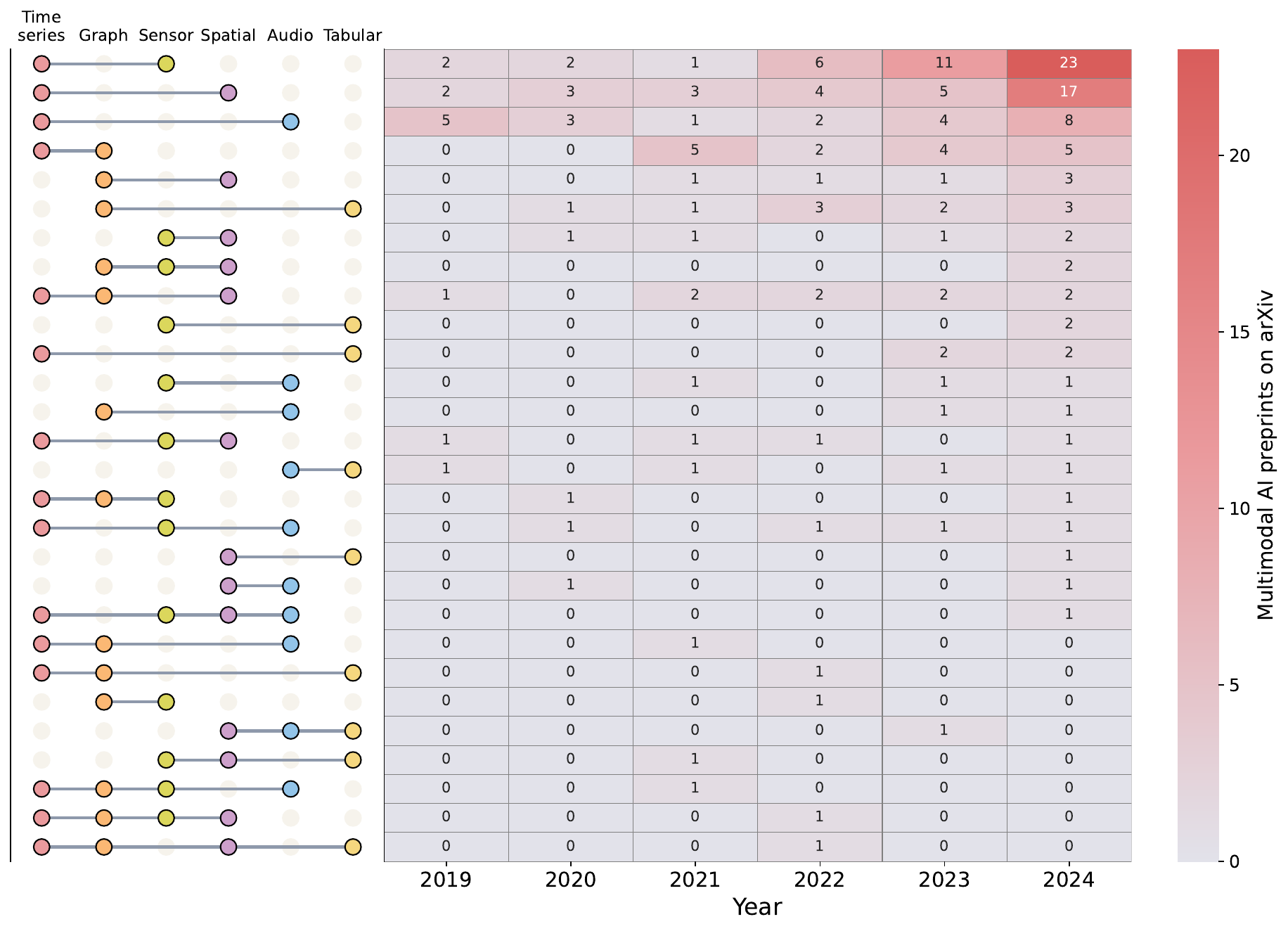}
    \caption{\textbf{{\color{red}Underexplored modality combinations in multimodal AI (2019--2024).}} 
    {\color{myblue}
    Heatmap of combinations of non-vision, non-language modalities in multimodal AI preprints by year. Each row represents a modality combination, and each column corresponds to a publication year. Darker shades indicate higher counts, with rows ordered by their 2024 totals. The coloured circles on the left identify the modality combinations. For example, in 2023, only two preprints used graph and tabular data (sixth row, fifth column). Time series and sensor data form the most common combination, likely because sensor data are often recorded as time series. Time series and spatial data are the second most common, possibly due to the importance of spatiotemporal modelling. In contrast, combinations involving graph, audio, and tabular data remain sparsely studied. These gaps highlight untapped potential for multimodal AI beyond vision and language. See Supplementary Section S2 for details on data processing and modality extraction.}}              
\label{fig:underexplored_modality}
\end{figure*}

While multimodal AI receives increasing attention and holds great promise, {\color{myblue}research has primarily focused on vision-language models \cite{bordes2024introduction,zhang2024vision} (Fig. \ref{fig:mmai-trend} and Supplementary Fig. \ref{fig:modality_combinations}), leaving other modalities--such as tabular and time-series data--and related disciplines underexplored \cite{vanposition} (Fig. \ref{fig:underexplored_modality})}. Real-world challenges, such as pandemic response, call for AI capable of integrating diverse types of data through {\color{myblue}interdisciplinary} collaboration, bridging the gap between research and application.

{\color{red}As modality integration broadens, multimodal-AI-specific deployment challenges increase, including missing modality \cite{ma2021smil, wu2024deep}, cross-modal alignment \cite{wang2022multi,liang2024foundations}, and multimodal privacy risk \cite{zhao2024survey,pranjal2023toward}. These issues are compounded by broader barriers such as data limitations, integration complexity, and domain-specific constraints \cite{paleyes2022challenges}. For example, 
rural healthcare may face limited compute infrastructure; financial services may face regulatory delays; and real-time applications such as autonomous driving demand strict latency control. Multimodal setups often amplify these challenges due to increased system complexity. Simply combining diverse modalities is not enough; real-world success requires proactive alignment with deployment constraints from the outset.}

This Perspective outlines a deployment-centric framework for multimodal AI that incorporates deployment constraints early and addresses challenges specific to multimodal integration across disciplines. We first present a general workflow for developing deployment-ready multimodal AI systems. We then examine three data-intensive, cross-disciplinary use cases, pandemic response, self-driving car design, and climate change adaptation, to illustrate common barriers and actionable strategies. By exploring underrepresented modalities and fostering open, interdisciplinary collaboration, we aim to broaden the impact of multimodal AI beyond vision and language.

\wordcounttext{introduction:}

\begin{figure*}[!t]
\centering
    \includegraphics[width=0.9\textwidth]{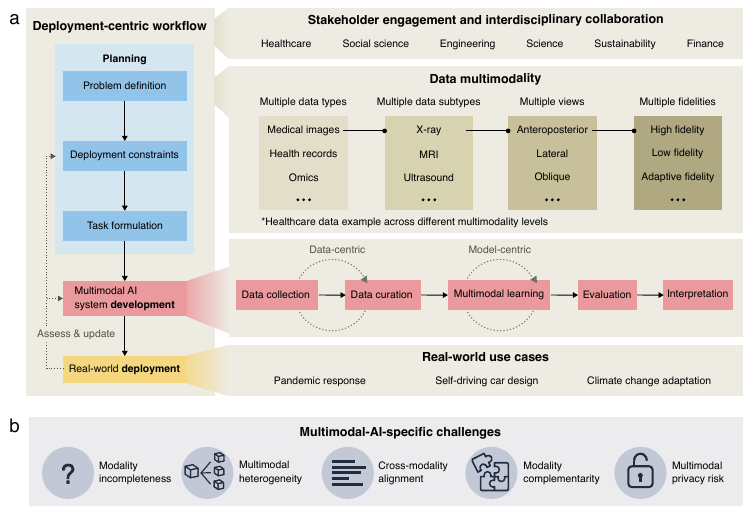}
    \caption{\textbf{Deployment-centric multimodal AI: workflow and challenges.} \textbf{a,} Deployment-centric multimodal AI workflow designed to meet real-world needs through a structured three-stage process covering planning, development, and deployment, with iterative assessments and updates. In particular, the planning stage considers deployment constraints early to ensure alignment with real-world settings and practical needs. This workflow incorporates {\color{myblue}stakeholder engagement and interdisciplinary collaboration at all stages to ensure that real-world needs and discipline-specific knowledge inform and enhance AI system development}. Moreover, we consider multiple levels of data multimodality, from data types and subtypes to views and fidelities, as illustrated with a healthcare example. This broader definition of multimodality offers new perspectives and rich options for leveraging the benefits of multimodality. The system development stage has five steps similar to a standard machine learning pipeline, where the data-centric and model-centric approaches \cite{seedat2023navigating} are indicated in the figure to highlight their differences from the deployment-centric approach. \textbf{b,} Five multimodal-AI-specific challenges shared across multiple disciplines and real-world applications{\color{red}: modality incompleteness, where one or more modalities are missing at training or deployment; multimodal heterogeneity, reflecting incompatible formats and data structures; cross-modality alignment, which requires synchronising data in time or meaning; modality complementarity, where the goal is to maximise synergy without introducing redundancy; and multimodal privacy risk, where data fusion increases the chance of sensitive re-identification.}}
\label{fig:multimodal-ai-flow}
\end{figure*}

\section*{Deployment-centric multimodal AI system development}
Traditionally, multimodal AI research has been \textit{model-centric} \cite{baltruvsaitis2018multimodal,ektefaie2023multimodal,xu2023multimodal}, focusing on developing new models to outperform existing ones on standardised benchmarks and datasets. The rise of generative AI \cite{jo2023promise} models such as ChatGPT \cite{brown2020language,achiam2023gpt} resulted in a shift towards \textit{data-centric} approaches \cite{seedat2023navigating,liu2024visual,li2024llava}, emphasising data resources and quality for better performance. However, the gap between high expectations and limited real-world impacts \cite{roberts2021common, paleyes2022challenges} indicates a pressing need for \textit{deployment-centric} approaches that prioritise real-world applicability, user needs, and ethical considerations, ensuring that AI innovations are both novel and practical, with a positive impact. We advocate a deployment-centric workflow for advancing multimodal AI from research to scalable solutions, built on ideas from machine learning operations lifecycle guidelines \cite{kreuzberger2023machine} and technology readiness levels for machine learning systems \cite{lavin2022technology}.

Figure \ref{fig:multimodal-ai-flow}a illustrates our deployment-centric workflow in three stages: planning, development, and deployment. Planning focuses on defining the problem, determining the suitability of multimodal AI over unimodal AI, understanding real-world constraints, and formulating AI tasks grounded in realistic assumptions. Development builds the multimodal AI system to learn predictive models from multimodal data, and deployment assesses real-world performance, feeding back insights to improve earlier steps. {\color{myblue}Guidance from stakeholder engagement and interdisciplinary collaboration (the top right of Fig. \ref{fig:multimodal-ai-flow}a) informs the entire workflow, ensuring that the perspectives of domain experts, end-users, and decision-makers are integrated for a robust and socially aligned system \cite{nielsen2025intersectional,lekadir2025future}.}

\subsection*{Planning for multimodal AI systems}
Planning (the three blocks on the top left of Fig. \ref{fig:multimodal-ai-flow}a) begins with \textbf{problem definition}, which involves clearly articulating the problem’s scope and objectives, and preliminarily assessing whether incorporating multimodal data can offer meaningful advantages, such as improved predictive performance or deeper insights, over unimodal approaches \cite{huang2021makes}. Beyond vision (image, video) and language (text), key modalities include audio, numeric time series (e.g. financial data sequences), sensor signals (e.g. wearable physiological measurements), and spatial (geolocation), tabular (structured data such as clinical records), and graph-based (relationships or networks) data. Furthermore, multimodality exists at multiple levels: multiple data types, subtypes, views, and fidelities \cite{meng2021multi,Penwarden2022MultifidelityPINNs} (the middle right of Fig. \ref{fig:multimodal-ai-flow}a {\color{myblue} and Fig. \ref{fig:dataset}}). For example, in healthcare, multimodal data can range from different types, e.g. medical images, electronic health records (EHRs), and omics data \cite{zitnik2019machine,lunke2023integrated}, to subtypes within a data type, e.g. X-ray, magnetic resonance imaging (MRI), and ultrasound within the imaging modality, multiple views of the same data (sub)type, e.g. anteroposterior, lateral, and oblique views of X-ray, and multiple fidelities, e.g. high, low, and adaptive fidelities of anteroposterior X-ray. Under this broad definition, the potential number of modalities can expand substantially as more levels are considered. Understanding these modalities early helps select those suited to the problem’s complexity and capable of offering insights beyond unimodal data.

After establishing the potential benefits of multimodality, we move on to \textbf{deployment constraints}, which involves examining the AI system’s application context, including user needs, data availability, regulatory compliance \cite{wu2021medical}, ethical considerations, societal impact, and economic trade-offs between high-cost and low-cost data modalities. This step ensures that the chosen modalities and the resulting multimodal AI system are not only technically feasible but also viable within the intended deployment environments. Understanding these constraints early informs task formulation, helping align solutions with real-world requirements and making it more effective and responsible.

The final step, \textbf{task formulation}, translates the defined problem and constraints into specific AI tasks, specifying the inputs, outputs, and evaluation metrics required to meet the objectives set during problem definition. This provides a clear development roadmap, ensuring multimodal AI systems are well-positioned to meet both technical and practical needs. {\color{myblue}Selecting modalities requires balancing utility, acquisition cost, complexity, and deployment feasibility. Fewer well-curated modalities may outperform broader but less practical combinations.}

\subsection*{Multimodal AI system development}
Developing multimodal AI systems (the lower left of Fig. \ref{fig:multimodal-ai-flow}a) parallels standard machine learning system development but introduces added complexities due to the integration of diverse data modalities. We have identified five multimodal-AI-specific challenges (Fig. \ref{fig:multimodal-ai-flow}b). \textbf{Modality incompleteness} occurs when one or more modalities are missing during training or deployment, necessitating the development of highly flexible or generative models.  \textbf{Multimodal heterogeneity} deals with varying formats, scales, and data structures, necessitating careful integration strategies to ensure interoperability. \textbf{Cross-modality alignment} ensures that the timing or meaning of data from different sources is correctly aligned for coherency and consistency, whether temporal (e.g. matching timestamps) or semantic (e.g. identifying data representing the same concept across modalities). \textbf{Modality complementarity} ensures that different modalities contribute complementary information to enhance performance, as more modalities may not improve results, e.g. if they add noise or redundancy. Finally, \textbf{multimodal privacy risk} \cite{zhao2024survey, pranjal2023toward} arises when independently anonymised datasets become identifiable again (re-identification) through their fusion, thereby revealing sensitive information and necessitating robust privacy-preserving techniques.

The development process consists of five key steps: data collection, data curation, multimodal learning, evaluation, and interpretation (the lower right of Fig. \ref{fig:multimodal-ai-flow}a). Each step plays a key role in addressing the multimodal-AI-specific challenges described above.

\textbf{Data collection} and \textbf{curation} build on the modality choices identified during planning, gathering relevant data from multiple sources and preparing them for AI model training. Diverse and reliable data sources ensure a high-quality, comprehensive representation of the problem space. {\color{myblue}Synthetic data and weak supervision provide valuable alternatives when labelled data is scarce \cite{van2024synthetic}, such as in rare diseases or time-sensitive crises, where expert annotation may be infeasible.} Once collected, the data undergoes standard curation, such as wrangling, cleaning, annotation, handling missing values, and quality assurance \cite{paleyes2022challenges}, tailored to multimodal AI. For instance, in healthcare, linking multimodal data from EHRs, imaging, and biosignals poses substantial challenges due to privacy concerns \cite{zhao2024survey} and the need for standardisation. The integration and cross-referencing of these modalities can amplify the risk of re-identification, underscoring the importance of addressing multimodal privacy risk through robust privacy-preserving techniques \cite{al2019privacy, Wendland:2022vo, che2023MultimodalFL}. Moreover, selected modalities should be interoperable and complementary, with their alignment and integration resulting in a cohesive, high-quality dataset suited to technical and deployment requirements.

\textbf{Multimodal learning} integrates diverse data modalities to leverage their unique strengths for improved predictive performance \cite{baltruvsaitis2018multimodal}. Traditional fusion strategies, early fusion (combining raw features), intermediate fusion (merging processed features), and late fusion (integrating outputs from independently processed modalities), offer trade-offs between model complexity and when modalities interact. Recent advances, hybrid fusion \cite{Zhao2024DeepMD} and knowledge distillation \cite{Wang2020MultimodalLW}, provide greater flexibility, enabling the combination of multiple strategies or the transfer of knowledge from complex to simpler models. Techniques such as co-attention mechanisms \cite{Yu2017MultimodalFB} can further enhance integration and performance by dynamically adapting to cross-modal interactions. The choice of strategy depends on the specific problem and data characteristics, balancing integration benefits with the added complexity of managing multiple modalities.

\textbf{Evaluation} and \textbf{interpretation} support the reliability, effectiveness, and fairness of multimodal AI systems, especially in complex, high-stakes environments. Beyond standard performance metrics such as accuracy, evaluation should assess the contribution of individual modalities, the alignment and synergy between them, and the system's robustness to noise and variability. For example, evaluating safety in multimodal models \cite{weidinger2024holistic} is key for understanding how biases across modalities may interact or amplify one another \cite{luccioni2024stable}, potentially undermining system reliability. Interpretation \cite{murdoch2019definitions, rudin2019stop} ensures that model decisions are understandable and transparent. Techniques such as heat maps, $t$-SNE, and decision trees help visualise and explain how different modalities interact and contribute to outcomes, fostering user trust and supporting error analysis and model refinement. {\color{myblue}Uncertainty estimation \cite{gawlikowski2023survey}, via ensembling, calibration, or Bayesian approaches, supports more reliable decisions in high-stakes applications.}

\subsection*{Real-world deployment of multimodal AI systems} 
Deploying multimodal AI systems (the bottom of Fig. \ref{fig:multimodal-ai-flow}a) requires infrastructure for hosting, scaling, and ongoing management, with continuous monitoring to ensure the system remains effective and relevant \cite{paleyes2022challenges}. {\color{myblue}Infrastructural and regulatory conditions, including compute availability, connectivity, and legal safeguards, can determine what is practically feasible and where deployment may succeed or fail.} Unlike unimodal AI, these systems face unique challenges, such as integrating diverse data modalities in real-time environments and maintaining performance despite modality-specific issues, such as sensor failures or data stream interruptions. Robust monitoring mechanisms enable the system to detect and compensate for failures in one modality by leveraging information from others. Additionally, real-time multimodal data fusion, where multimodal data may arrive asynchronously, requires careful infrastructure design and adaptive algorithms.

The following sections examine three cross-disciplinary use cases that demonstrate the value and applicability of the deployment-centric multimodal AI workflow: pandemic response, self-driving car design, and climate change adaptation.

\wordcounttext{technical overview:}

\section*{Use case 1: Pandemic response}

\begin{figure*}[!ttt]
\centering
    \includegraphics[width=.95\textwidth]{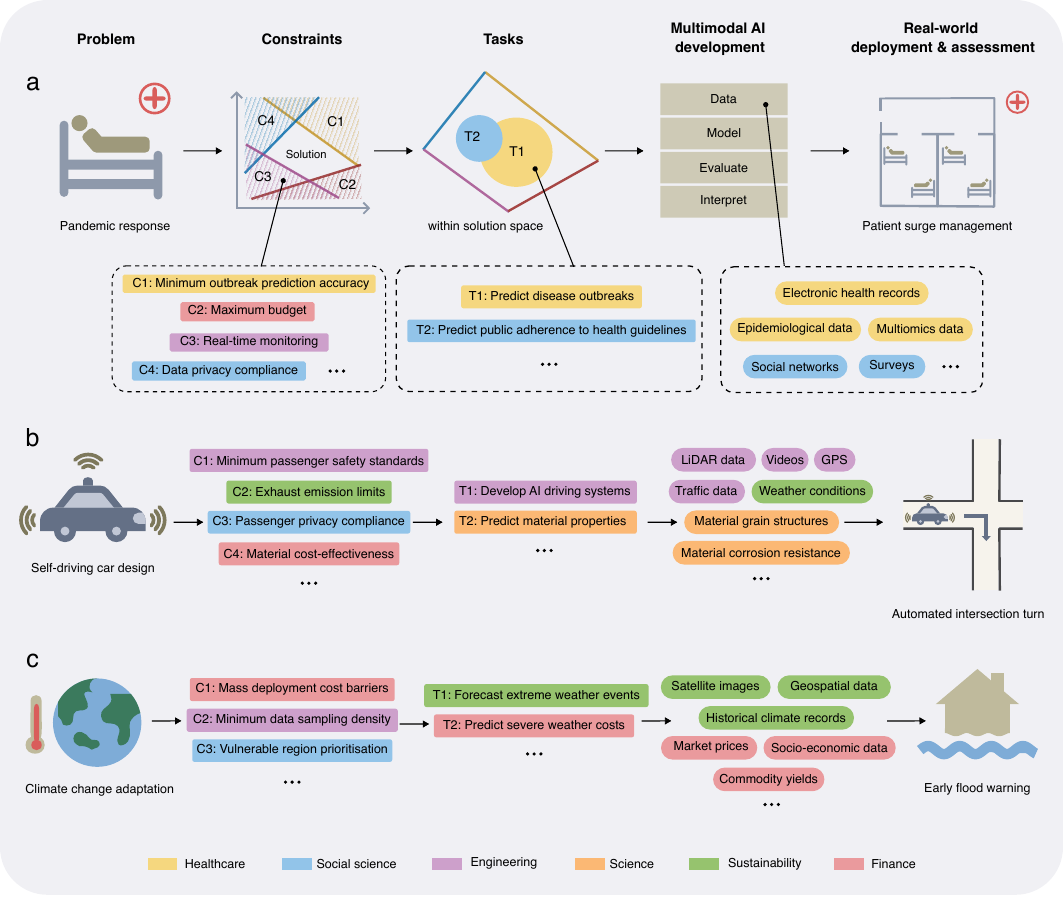}
    \caption{\textbf{Deployment-centric multimodal AI workflow for {\color{myblue}three use cases}.} {\color{myblue}These examples demonstrate the versatility of the proposed workflow, showing how deployment-centric considerations can be applied across various disciplines to solve complex challenges.} \textbf{a,} This example illustrates how the proposed deployment-centric multimodal AI workflow can better address real-world challenges in pandemic response, drawing on {\color{myblue}stakeholder engagement and interdisciplinary} collaboration, as indicated by the different colours across stages. The process begins with defining the problem of interest properly and carefully considering the space of various deployment constraints in order to formulate specific tasks. Two tasks are shown in the figure: one with a focus on healthcare and the other on social science. Next, we develop multimodal AI systems using diverse data sources. Finally, we proceed to real-world deployment and assessment, exemplified by the application of patient surge management. {\color{myblue}\textbf{b,} In self-driving car design, the workflow incorporates deployment constraints from social science, engineering, sustainability, and finance disciplines. Two key tasks focus on engineering and science considerations, respectively. Multimodal AI development uses multiple data sources, including remote sensing, weather, and materials data, aiming for real-world deployment exemplified by automated intersection turning. \textbf{c,} In climate change adaptation, the workflow considers constraints such as mass deployment cost barriers, minimum data sampling density, and prioritisation of vulnerable regions. Task formulation and multimodal AI development focus on environmental sustainability and financial considerations, reflecting how task and model development are shaped by discipline-specific needs. Early flood warning serves as an example of real-world deployment and assessment.}}

\label{fig:mmai_use_case}
\end{figure*}
    
Pandemic response (Fig. \ref{fig:mmai_use_case}a) presents a complex challenge requiring coordinated efforts across healthcare and other disciplines \cite{khojasteh2022climate}. Defining the problem involves assessing early how multimodal AI can offer advantages over unimodal approaches by integrating diverse data sources for deeper insights and more accurate predictions. For example, in social science, combining textual content from surveys or social networks with tabular demographic data can help understand mental health impacts and detect changes in online behaviour \cite{elmer2020students,valdez2020social}. In sustainability, analysing time-series environmental data, satellite imagery, and sustainability reports can reveal the environmental effects of pandemics \cite{liu2020near,zheng2020satellite}. In finance, integrating economic indicators, transaction patterns, and market response data with health data can predict socio-economic impacts and inform public health strategies \cite{mosser2020central,nicola2020socio}. A well-defined problem ensures clear constraints and tasks can be established.

The deployment of multimodal AI in pandemic response faces privacy, technical, and economic constraints. Privacy concerns arise from integrating diverse data sources such as EHRs, social media, and genomic information, which increases the risk of re-identification and amplifies the potential impact of data breaches. Maintaining public trust requires robust privacy-preserving techniques \cite{al2019privacy, Wendland:2022vo, che2023MultimodalFL} and compliance with regulatory frameworks. Technical challenges include real-time monitoring, where multimodal AI systems must integrate and process data streams rapidly for timely interventions. The heterogeneity of health and social data complicates alignment and standardisation, particularly in international or multi-centre collaborations. Achieving minimum predictive accuracy thresholds for outbreak detection and public health decision-making helps build trust in AI-driven systems. Economic constraints, including computational costs and budget limitations, restrict accessibility and scalability. High-cost modalities (e.g. MRI) may be impractical in low-resource settings, so multimodal AI systems should flexibly support lower-cost alternatives (e.g. wearables) to promote equitable deployment.

Task formulation translates the defined problem into specific AI objectives, detailing inputs, outputs, and evaluation metrics for pandemic-related challenges. A primary task is predicting disease outbreaks by integrating multimodal data such as epidemiological records, EHRs, and environmental factors to inform early interventions and resource allocation. Another task is monitoring public adherence to health guidelines using social media data, geospatial information, and biosignals, providing actionable insights to refine policies. Additionally, multimodal AI can support real-time decision-making in healthcare by combining patient data streams, including wearable biosignals, medical images, and clinical notes, for managing patient surges effectively \cite{ding2020wearable, charlton_wearable_2022}. Task formulation ensures that multimodal AI systems align with technical and practical needs, strengthening pandemic response through actionable insights and tailored interventions.

Developing multimodal AI systems for pandemic response follows the standard workflow of the five key steps outlined above. Data collection assembles diverse sources, such as EHRs, genomic data, and biosignals, to create comprehensive datasets for outbreak prediction and patient management. During curation, standardisation and alignment address challenges such as integrating time-sensitive data streams from biosignals and geospatial sources. Multimodal learning can employ self-supervised \cite{zong2024self} and cross-domain \cite{yang2014cross} techniques to capture cross-modal relationships effectively and integrate domain-specific knowledge. Evaluation requires multi-metric, multi-centre validation \cite{han2024randomised, plana2022randomized}, ensuring robustness for tasks such as outbreak prediction and public adherence monitoring. Interpretability and explainability enable healthcare professionals and policymakers to trust and act on multimodal AI insights in high-stakes scenarios \cite{imrie2023multiple}. Tailoring these development steps to pandemic-specific needs ensures adaptability, reliability, and ethical compliance in rapidly changing contexts.

Deploying multimodal AI systems in pandemic response translates research into actionable solutions for high-stakes environments. For instance, patient surge management can integrate biosignals, medical images, and clinical notes to optimise intensive care unit bed allocation and staff deployment. Public health decision-making can benefit from multimodal analyses of epidemiological data, EHRs, and social media trends to guide interventions such as lockdowns or resource distribution. Real-time deployment requires robust infrastructure to process asynchronous data streams and address modality-specific interruptions, such as gaps in biosignal or social media data. Adaptive algorithms and reliable systems ensure timely and accurate outbreak prediction and monitoring. Assessment through multi-centre trials and multi-disciplinary benchmarks \cite{mincu2022developing} can validate the added value of multimodal AI over unimodal systems.
Measuring the accuracy of outbreak prediction or the effectiveness of intervention informed by multimodal AI can ensure trust and accountability. Further addressing scalability and privacy compliance can support sustainable and ethical deployment.

\wordcounttext{pandemic response:} %

\section*{Use case 2: Self-driving car design}
Self-driving car design (Fig. \ref{fig:mmai_use_case}b) exemplifies a transformative application of multimodal AI, addressing challenges in safety, efficiency, and sustainability \cite{soni2021design,badue2021self}. Defining the problem involves determining how multimodal AI can substantially outperform unimodal approaches for navigating complex environments safely and efficiently. For instance, fusing sensors such as LiDAR (light detection and ranging), radar and cameras can enhance perception and decision-making in dynamic settings, especially where a single sensor may fail \cite{yeong2021sensor}. Beyond perception, multimodal AI can support physical simulations, such as combining text, images, and videos, to model vehicle dynamics and road interactions \cite{yang2024generalized}. In materials design, integrating structural and chemical data aids in developing lightweight, sustainable materials for vehicle construction \cite{zhang2022advanced}. Clearly defining these challenges ensures multimodal AI solutions are effectively tailored to the demands of autonomous driving.

The deployment of multimodal AI in self-driving car design faces safety, privacy, technical, and economic constraints that must be addressed for safe and scalable adoption. Privacy concerns arise from integrating multimodal data, such as LiDAR, cameras and passenger-related information, which increases the potential for data misuse or unauthorised tracking \cite{hansson2021self}. Mitigating these risks requires robust data anonymisation and adherence to privacy regulations \cite{chowdhury2020attacks}. Technical challenges include real-time data processing in dynamic environments, such as urban intersections or adverse weather. Modality-specific failures, such as LiDAR disruptions in heavy rain, require fallback mechanisms, such as radar to maintain reliability. 
Self-driving systems must achieve high predictive accuracy for collision avoidance and route planning, and operate reliably across regions with varying regulations and infrastructure. Economic constraints involve balancing the cost of high-resolution sensors with scalability needs. Meeting sustainability goals requires energy-efficient operation, with trade-offs between sensor performance and power consumption. Addressing these constraints holistically enables robust and accessible deployment of self-driving technologies.

In self-driving car design, task formulation maps problems and constraints to concrete AI objectives for perception, navigation, and safety. A key task is developing AI driving systems that integrate multimodal data streams, including LiDAR, radar, GPS, and cameras, for real-time navigation. These systems must dynamically process inputs to handle traffic, obstacles, and weather variability. For example, integrating road and weather data can enhance route planning and safety in adverse conditions \cite{dey2014potential}. Another task is predicting material properties and performance to support vehicle design. By fusing structural, experimental, and supply chain data, multimodal AI can optimise materials for lightweight, durable, and cost-effective vehicles, aligning with sustainability goals \cite{kamran2022artificial}. Additional tasks include passenger-focused objectives, such as integrating biosignals and cabin sensors to enhance comfort and safety. Ensuring privacy compliance remains integral across all tasks.

Developing multimodal AI systems for self-driving cars follows the five-step workflow above, adapted to the demands of autonomous systems. Data collection involves assembling sensor inputs (e.g. LiDAR, radar, cameras) and materials data to build comprehensive datasets. Data curation addresses challenges such as synchronising asynchronous sensor inputs and standardising materials information. Learning can employ fusion strategies to enhance system reliability, while physics-informed neural networks \cite{raissi2019physics, karniadakis2021physics} can integrate scientific laws to improve realistic simulations of driving scenarios and materials development. Evaluation ensures robustness through multi-metric validation across diverse scenarios, including urban environments and extreme weather conditions. Finally, interpretation in autonomous systems supports real-time explainability and transparency, enabling stakeholders to trust and audit system decisions. By aligning these steps with the unique requirements of autonomous systems, multimodal AI can ensure adaptability, safety, and performance.

Real-world deployment of multimodal AI systems for self-driving cars requires live sensor integration, infrastructure readiness, and operational resilience across diverse environments. Robust in-vehicle computing and inter-vehicle communication enable reliable and coordinated performance at scale. Assessment validates multimodal AI’s added value through simulations and multi-centre trials. Testing performance under extreme conditions, such as adverse weather or high-traffic intersections, ensures reliability. Addressing scalability and privacy compliance further supports ethical, sustainable deployment, advancing societal trust in self-driving technologies \cite{liu2019public}.

\wordcounttext{self-driving car design:}

\section*{Use case 3: Climate change adaptation}
Climate change adaptation (Fig. \ref{fig:mmai_use_case}c) involves forecasting extreme weather, assessing climate risks, and managing resources to minimise environmental and socio-economic impacts \cite{bi_accurate_2023,lam2023learning,mathiesen2018rating,breitenstein2022disclosure}. Defining the problem involves assessing whether multimodal AI can surpass unimodal approaches by integrating diverse data sources for deeper insights and actionable strategies. For example, multimodal AI can combine satellite imagery, historical climate records, and geospatial data to model weather patterns and predict extreme events, supporting proactive disaster management \cite{imran2020using}. Integrating socio-economic data with environmental observations can enable more comprehensive climate risk assessments \cite{tanir2024social}.

Deploying multimodal AI for climate change adaptation must navigate privacy, technical, and economic constraints. Privacy concerns arise from integrating sensitive data, such as socio-economic records and proprietary satellite imagery. Ensuring compliance with global and regional privacy regulations and the ethical use of data supports responsible deployment. Technical challenges include the heterogeneity and sparsity of environmental and socio-economic data. Economic constraints, such as the high cost of advanced sensors, necessitate balancing low-cost and low-fidelity data sources with high-cost, high-fidelity alternatives to maintain accessibility and scalability. By addressing these constraints, multimodal AI can deliver equitable and sustainable climate solutions in diverse global contexts.

For climate change adaptation, task formulation aligns complex environmental challenges with actionable AI tasks. A primary task is forecasting extreme weather events by integrating multimodal data such as satellite imagery, historical climate records, and real-time observations. These forecasts can enable timely interventions to mitigate human and economic losses \cite{kumar2021overview},  including early warning systems for floods that integrate geospatial, hydrological, and social data streams \cite{tkachenko2017predicting}. Another task is predicting the socio-economic impacts of severe weather by combining spatial, market, and socio-economic data, guiding resource allocation and policy decisions \cite{thulke2024climategpt}. These tasks align multimodal AI with both technical requirements and societal priorities, advancing climate resilience and equitable adaptation strategies.

Developing multimodal AI systems for climate change adaptation follows a similar five-step workflow. Data collection gathers remote sensing data, near-real-time sensor network outputs, historical climate records, and socio-economic metrics, creating comprehensive datasets for tasks such as extreme weather forecasting and early warning systems. Data curation addresses challenges such as temporal misalignment and data sparsity by standardising inputs and enriching sparse data through interpolation or integration of auxiliary sources. Multimodal learning can benefit from advanced models such as Aurora \cite{bodnar2024aurora}, which fuse satellite imagery and meteorological data to improve atmospheric predictions. Evaluation uses multi-metric benchmarks to test system performance in diverse scenarios. For example, WeatherBench 2 provides a standard platform for assessing atmospheric prediction models \cite{rasp2024weatherbench}. Interpretation ensures outputs are transparent and actionable for stakeholders, supporting evidence-based decision-making. By aligning these steps with climate-specific challenges, multimodal AI can deliver reliable, scalable, and adaptable solutions.

Real-world deployment of multimodal AI for climate change adaptation translates research insights into scalable, operational systems. Early warning platforms and impact forecasting models must synchronise diverse data streams, deliver real-time outputs, and integrate with decision-making infrastructures across sectors and regions \cite{morshed2024decoding,Reichstein2025EarlyWO}. Real-time monitoring requires robust infrastructure to synchronise and process diverse modalities. Deployment challenges include latency, data coverage gaps, and variable infrastructure capacity across geographies. Assessment involves stress testing and multi-centre validation to ensure reliability across environmental and socio-economic scenarios. Ethical considerations, such as prioritising vulnerable communities and ensuring equitable access to data and AI tools, help enable sustainable deployment. By overcoming deployment challenges, multimodal AI systems can empower policymakers, industries, and communities to respond proactively to climate challenges, advancing global sustainability efforts.

\wordcounttext{climate change adaptation:} %

\section*{Outlook}

Most existing research on multimodal AI has focused on model-centric development. While attention to data-centric development is increasing, unlocking the full potential of multimodal AI and addressing real-world challenges will ultimately require a shift towards deployment-centric development. This shift will require strategic advancements in data, model, and deployment methods, as well as concerted efforts to address multimodal-AI-specific challenges, namely modality incompleteness, multimodal heterogeneity, cross-modality alignment, modality complementarity, and multimodal privacy risk, through {\color{myblue}stakeholder engagement,} interdisciplinary collaborations, and community-building. Box~1 presents strategic recommendations to advance multimodal AI across disciplines under a deployment-centric perspective. These recommendations draw on insights from the three use cases and the broader cross-disciplinary analysis in the Supplementary Information on multimodal AI beyond these three use cases, providing actionable guidance for future developments.

Deployment-centric development brings challenges around safety, reliability, interpretability, scalability, and ethics, particularly as multimodal AI expands in high-stakes applications such as healthcare and sustainability. Addressing these challenges requires implementing robust human-in-the-loop systems, developing clear standards for safety and transparency, and prioritising scalable, resource-efficient infrastructure to support increasing demands for data integration, processing, and storage.

\begin{figure*}[!t]
\centering
    \includegraphics[width=.98\textwidth]{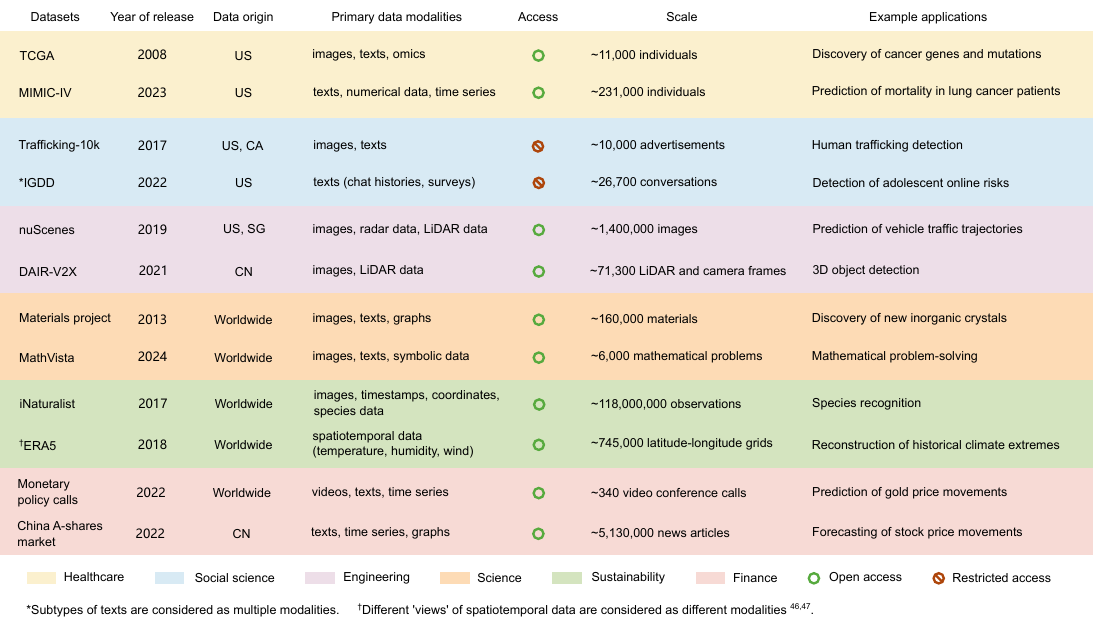}
    \caption{\textbf{Examples of multimodal benchmark datasets.} We selected two illustrative examples for each of the six disciplines to showcase diverse multimodal data and provide a starting point for multimodal AI exploration. For each dataset, we report six key attributes: year of release, data origin, data modalities, accessibility, scale of the primary modalities, and example applications. In each discipline, datasets are listed in ascending order by year of release. For the healthcare, science, and finance disciplines, we selected one small-scale dataset for quick experimentation and one larger-scale dataset for comprehensive exploration. The datasets include:
    TCGA (The Cancer Genome Atlas) \cite{tomczak2015review};
    MIMIC-IV (The Medical Information Mart for Intensive Care IV database) \cite{johnson2023mimic};
    Trafficking-10k (human trafficking advertisement dataset) \cite{tong2017combating}; 
    IGDD project (Instagram data donation project) \cite{razi2022instagram}; 
    nuScenes (autonomous driving scene dataset) \cite{caesar2020nuscenes}; 
    DAIR-V2X (real-scenarios vehicle to everything dataset) \cite{yu2022dair}; 
    Materials project (material property dataset) \cite{jain2013commentary};
    MathVista (mathematical reasoning dataset) \cite{lu2024mathvista};
    iNaturalist (citizen science platform for biodiversity data) \cite{inaturalist};
    ERA5 (European Centre for Medium-Range Weather Forecasts Reanalysis v5) \cite{hersbach2020era5};
    Monetary policy calls (monetary policy call dataset) \cite{mathur2022monopoly};
    China A-shares market (dataset of public companies listed in China A-shares market) \cite{cheng2022financial}. 
    {\color{myblue}
    Modality definitions can vary. IGDD and ERA5 are not strictly multimodal in the traditional sense, but we treat them as such under the broader definition presented in this Perspective. The IGDD dataset contains only text data, but we consider its subtypes (chat histories and surveys) as distinct text modalities. Similarly, ERA5 comprises spatiotemporal variables such as temperature, humidity, and wind, which we treat as complementary `views' offering distinct information streams for climate modelling, as adopted in prior work \cite{chen2023fengwu, jin2023spatiotemporal}.}
    }
\label{fig:dataset}
\end{figure*}

Robust data-centric development underpins deployment-centric progress by ensuring data availability, diversity, and quality. High-quality multimodal data is often limited, and existing datasets often lack the diversity needed to ensure fairness in AI systems. Clear benchmarks \cite{liang2021multibench,de2025towards} and globally accessible datasets {\color{myblue}(Fig. \ref{fig:dataset})} are needed to enhance trust, consistency, and adaptability across disciplines, facilitating reproducible research. Promoting secure data-sharing frameworks \cite{torabi2023common}, open initiatives such as the European Life Science Infrastructure for Biological Information \cite{crosswell2012elixir}, and globally accessible multimodal data platforms will further strengthen AI's capacity to address diverse global challenges. Knowledge graphs \cite{liang2024survey} also offer promise for organising varied data formats to unify and contextualise multimodal inputs across disciplines.

Model-centric development for multimodal AI faces unique challenges in effectively and efficiently fusing diverse modalities. While foundation models have proven effective for vision and language tasks, expanding them to other modalities and disciplines \cite{archit2025segment} could substantially lower development barriers. Guidelines and frameworks for selecting relevant modalities, comparing multimodal versus unimodal performance, and evaluating the benefits of using many versus few modalities will help maximise model efficiency and utility. {\color{red}
As large language models, multimodal foundation models, and generative AI systems grow in prominence \cite{li2024multimodal,fei2022towards,narayanswamyscaling, cui2025towards}, deployment-centric design plays an increasingly important role in guiding architectural choices, managing inference-time costs, and aligning with data and domain-specific constraints \cite{anthropic2024mcp, weisz2024design}. These models also increasingly rely on synthetic or weakly labelled data, or cross-modal supervision to reduce annotation costs \cite{li2022blip, li2023blip, driess2023palm, tan2024large, zong2024self, ding2024data}, reinforcing the need for deployment-aware data strategies.}

{\color{myblue}Stakeholder engagement and interdisciplinary collaboration are both crucial for progress, as illustrated across the three use cases. 
Integrating stakeholder input ensures contextual relevance and trust. Standardising data practices, fostering cross-disciplinary knowledge exchange, and developing shared platforms help bridge disciplinary gaps and enable multimodal AI to address complex societal challenges more effectively and holistically~\cite{nielsen2025intersectional,lekadir2025future}.}

Building a dynamic and inclusive multimodal AI community fosters innovation and drives solutions to complex real-world challenges. Collaborative efforts, through regular workshops, forums, and interdisciplinary research initiatives, facilitate knowledge exchange and inspire collective problem-solving across disciplines. Engaging researchers, practitioners, and stakeholders from diverse backgrounds, particularly early-career researchers and underrepresented groups, will not only enrich the AI ecosystem but also broaden the perspectives and expertise shaping it, making it more `multimodal'. Community-driven initiatives \cite{Chan2020ReportingGF,de2022guidelines}, including comprehensive surveys or perspective papers, provide insights that guide future developments, ensuring multimodal AI advances in responsible, impactful directions.

Ultimately, the success of multimodal AI lies in its deployment. By embracing a deployment-centric mindset, the community can turn research breakthroughs into real-world impact across disciplines.

\newpage

\begin{tcolorbox}[naturebox]
\label{box:recommendations}
\begin{minipage}{0.49\textwidth}

\vspace{-2mm}

\textbf{Deployment-centric development}  
\begin{itemize}[left=1em]
            \item \textit{Safety, reliability, and interpretability}
                
                \textbf{Challenge}: Ensuring reliable, safe deployment across varied conditions and building user trust to facilitate understanding and broader adoption.
                
                \textbf{Recommendation}: Translate complex data and outputs into accessible formats (images, text, or speech); incorporate domain-specific knowledge; implement human-in-the-loop systems for verification and validation; conduct rigorous usability testing; develop standards for safety, reliability, and interpretability criteria; and enforce such criteria in research and peer review \cite{nielsen2025intersectional}.                                   
            \item \textit{Scalability and resource efficiency}
                
                    \textbf{Challenge}: Scaling multimodal AI to handle vast data volumes while optimising resource consumption, maintaining system availability, and ensuring uninterrupted communication between system components.
                    
                                        
                    \textbf{Recommendation}: Innovate scalable and resource-efficient multimodal AI solutions; promote collaboration between AI developers and hardware providers for sustainable resource use; ensure stable system availability and seamless inter-component communication; and develop cloud-based solutions and edge computing to power scalability.
                
            \item \textit{Ethical compliance and user preparedness}
                
                     \textbf{Challenge}: Addressing privacy, consent, and bias in applications with profound societal impacts.
                    
                     \textbf{Recommendation}: Develop clear ethical guidelines; include end-users in the design; involve human oversight and create user feedback loops for continuous solution refinement; and provide comprehensive training for reliable deployment.            
                
        \end{itemize}
\textbf{Data-centric development} 
        \begin{itemize} [left=1em]       
            \item \textit{Data scarcity and access}
                
                \textbf{Challenge}: 
                Limited availability of high-quality multimodal datasets spanning a comprehensive range of modalities, complicated further by privacy and ethical constraints and additional heterogeneity when data sharing spans secure data environments, organisations, and regions.
                
                \textbf{Recommendation}: 
                Promote open data initiatives, such as anonymised data-sharing programmes; invest in cross-discipline efforts for data collection and curation to improve availability; build data infrastructures that ensure secure, privacy-compliant, and ethics-compliant access across domains and regions; establish guidelines to standardise formats and annotations; and develop advanced tools such as knowledge graphs to integrate structured and unstructured data sources for improved contextual understanding, reasoning, and interoperability.                               
                
            \item \textit{Balanced data representation}
                 
                    \textbf{Challenge}: Bias due to the limited availability of data representing underrepresented or less-studied categories (e.g. populations and regions).
                    
                    \textbf{Recommendation}: Diversify dataset development to enhance model generalisability and reduce bias; and build and enhance multimodal data platforms (e.g. UK Biobank \cite{bycroft2018uk}, MIMIC \cite{johnson2016mimic}, Materials Project \cite{jain2013commentary}, ERA5 \cite{hersbach2020era5}) to improve global data accessibility and quality.                    
        \end{itemize}        
\end{minipage}
\hfill
\begin{minipage}{0.49\textwidth}

                 
                    

\textbf{{\color{myblue} Stakeholder engagement and interdisciplinary collaboration}}  
        \begin{itemize}[left=1em]
        {\color{myblue}
            \item \textit{Stakeholder inclusion and alignment}
                
                    \textbf{Challenge}: Limited involvement of stakeholders (e.g. domain experts, end-users, and regulators) during early-stage planning can result in solutions that are technically sound but poorly aligned with deployment contexts.                 
                    
                    \textbf{Recommendation}: Integrate stakeholder input across the AI lifecycle through co-design, participatory planning, and regulatory consultation to ensure contextual relevance, trust, and successful real-world adoption.
        }
            \item \textit{Cross-disciplinary standards and communication}
                
                    \textbf{Challenge}: Discrepancies in multimodal data standards and communication barriers across disciplines.
                    
                    \textbf{Recommendation}: 
                    Promote cross-disciplinary standards for multimodal data integration; and organise interdisciplinary exchange events to align goals, foster collaboration, and build a thriving ecosystem.                    
            \item \textit{Intellectual property (IP) and workflow adaptation}
                
                    \textbf{Challenge}: IP concerns and resistance to change established workflows.
                    
                    \textbf{Recommendation}: Develop adaptive strategies and open collaborative platforms to facilitate cross-disciplinary projects.            
        \end{itemize}    

\vspace{2.4cm}

\textbf{Model-centric development}
        \begin{itemize}[left=1em]
            \item \textit{Multimodal fusion and modality selection}
                     
                    \textbf{Challenge}: Managing diverse properties and patterns across modalities, including avoiding unnecessary modality redundancy, to maintain integrity and utility while enhancing integration.
                    
                    \textbf{Recommendation}:                     
                    Assess the value of additional modalities by comparing unimodal, few-modality, and full-modality models; enhance semantic modality alignment, e.g. via large language models; develop guidelines for optimal modality selection {\color{red}aligned with deployment context and cost-performance constraints}; and address other multimodal-AI-specific modelling challenges (Fig.~\ref{fig:multimodal-ai-flow}b).


            \item \textit{Foundation models (FMs)}
                 
                    \textbf{Challenge}: 
                    Developing FMs beyond vision and language requires substantial resources, limiting accessibility across disciplines.                    
                    
                    \textbf{Recommendation}: 
                    Develop and expand multimodal FMs beyond vision and language as critical reusable infrastructure to lower barriers; invest in downstream research and advance methods to reduce resource costs, {\color{red}e.g. using synthetic or weakly labelled data to ease training demands \cite{van2024synthetic}; and pair underrepresented modalities (e.g. graphs) with more interpretable ones (e.g. language) to improve usability and broaden accessibility for non-expert stakeholders}.
                    
        \end{itemize}        
\end{minipage}
\end{tcolorbox}

\wordcounttext{outlook:}

\section*{Data availability}
{\color{myblue}Source data for Fig. \ref{fig:mmai-trend}, Fig. \ref{fig:underexplored_modality}, and Supplementary Fig. \ref{fig:modality_combinations} are available with this paper and at \url{https://github.com/multimodalAI/multimodal-ai-landscape}, where they will be updated annually.}

\putbib
\end{bibunit}

\section*{Acknowledgements}
This work was enabled and supported by the Alan Turing Institute. We thank T. Chakraborty and C. Li for inspiring this work, D. A. Clifton for his invaluable support, and T. Dunstan for contributing to the climate change adaptation section. 
J.Z. is supported by donations from D. Naik and S. Naik.
S.Z. is supported by EPSRC (grant no. EP/Y017544/1).
T.L.vdP. was supported by EPSRC (grant no. EP/Y028880/1). 
A.V. is supported by UKRI CDT in AI for Healthcare (grant no. EP/S023283/1).
A.G. is supported by the Research Council of Norway (secureIT project, no. 288787).
M.Z. is supported by EPSRC (grant no. EP/X031276/1).
C.T. is supported by UKRI CDT in AI-enabled Healthcare (grant no. EP/S021612/1).
R.L. is supported by the Royal Society (grant no. IEC\textbackslash NSFC\textbackslash233558).
L.V.Z. is supported by NERC (grant no. NE/W004747/1).
O.T. is supported by UKRI CDT in Application of Artificial Intelligence to the study of Environmental Risks (grant no. EP/S022961/1).
Z.S. is supported by Google DeepMind.
O.A. is supported by NIHR Barts BRC (grant no. NIHR203330).
T.N.D. is supported by UKRI CDT in Accountable, Responsible and Transparent AI (grant no. EP/S023437/1).
L.F. is supported by MRC (grant no. MR/W006804/1).
N.J. is supported by the EU's co-funded HE project MuseIT (grant no. 101061441).
M.V. is supported by St George's Hospital Charity.
A.C-C. is supported by EPSRC (grant no. EP/Y028880/1).
H.W. is supported by MRC (grant no. MR/X030075/1).
C.C. is supported by the Royal Society (grant no. GS\textbackslash{R2}\textbackslash242355).
T.Z. was supported by the Royal Academy of Engineering (grant no. RF\textbackslash201819\textbackslash18\textbackslash109).
G.G.S. is supported by EPSRC (grant no. EP/Y009800/1).
P.H.C. is supported by BHF (grant no. FS/20/20/34626).
H.L. is supported by EPSRC (grant no. UKRI396).
The views expressed in this material are those of the authors and do not necessarily represent the views of their affiliated institutions or funders.

\section*{Author contributions}

X.L., J.Z., S.Z. and H.L. contributed equally.
X.L., J.Z., S.Z., T.L.vdP., S.T., T.Z., W.K.C., P.H.C., and H.L. conceptualised the manuscript. 
X.L., J.Z., S.Z., A.G., M.Z., L.vZ., O.T., and H.L. designed the figures.
X.L. and H.L. coordinated the entire project.
All authors contributed to writing, resources, or editing.

\section*{Competing interests}

G.G.S. is a scientific advisory board member at BioAIHealth. P.H.C. provides consulting services to Cambridge University Technical Services for wearable manufacturers. The remaining authors declare no competing interests.

\label{LastPageMain}

\clearpage
\setcounter{page}{1}
\pagestyle{suppstyle}  %
\resetlinenumber[1] %

\begin{bibunit}

\section*{\LARGE Supplementary information \\ \Large Towards deployment-centric multimodal AI beyond vision and language}

\section*{S1. Multimodal AI beyond the three use cases}

This section highlights discipline-specific advancements and challenges in multimodal AI that extend beyond the three primary use cases of pandemic response, self-driving car design, and climate change adaptation. By exploring advancements and challenges in healthcare, social science, engineering, science, sustainability, and finance, we showcase the versatility of multimodal AI across diverse domains. These insights underline the importance of interdisciplinary collaboration in addressing complex global challenges while paving the way for future advancements.

\subsection*{Healthcare}
Beyond its critical role in pandemic response, multimodal AI holds transformative potential in healthcare by advancing diagnostics, personalising treatment, and improving patient care. By integrating diverse data sources such as electronic health records (EHRs), imaging, biosignals, and multiomics, multimodal AI systems can provide comprehensive insights into patient health \cite{acosta2022multimodal}. For example, integrating wearable biosignals with EHRs can enable continuous monitoring, supporting the early detection of chronic conditions such as arrhythmias and heart failure \cite{krittanawong2021integration}. In oncology, the fusion of imaging modalities and multiomics data can enhance cancer diagnosis and treatment planning \cite{steyaert2023multimodal}, offering a more nuanced understanding of disease progression. Similarly, in neurology, combining imaging, sensor data, and clinical observations can facilitate the early detection of neurodegenerative diseases \cite{huang2023multimodal}.


Wider adoption of multimodal AI is hindered by challenges such as privacy concerns, the heterogeneity of healthcare data, and the scarcity of high-quality datasets. Addressing these challenges requires standardised methodologies, privacy-preserving techniques, and interdisciplinary collaborations. Building datasets that reflect the complexities of patient pathways and addressing bottlenecks in data integration are crucial steps towards integrating multimodal AI into routine clinical practice. Techniques such as federated learning, differential privacy, and privacy-preserving generative AI for synthetic multimodal data generation can help mitigate privacy risks, enabling the development of robust and trustworthy AI systems. Integrating multimodal data, such as multiomics \cite{lunke2023integrated}, into routine clinical workflows will require rigorous validation and improved standardisation. The limited use of randomised controlled trials (RCTs) to validate AI in clinical settings \cite{han2024randomised} has prompted the development of community-based guidelines \cite{Chan2020ReportingGF} to improve reliability and transparency.


  
\subsection*{Social science}
In social science, multimodal AI can drive innovations in behavioural analysis, public policy decision-making, and societal impact assessments. Beyond its applications in pandemic response, multimodal AI can help monitor societal behaviours and safeguard vulnerable populations, such as protecting children online, by integrating text, image and metadata from social media \cite{ali2023getting}. Multimodal AI systems can also help detect online criminal activity \cite{goyal2023detection} and analyse societal trends to inform public policy \cite{androutsopoulou2018framework}. Additionally, acoustic data can add depth to communication analysis by capturing emotional nuances \cite{zhu2024review}. Combining economic transaction data with geospatial information can reveal trends in societal decision-making \cite{Huang2024CrowdsourcedGD}.


Persistent challenges include restricted access to sensitive data such as social media or mental health records, privacy concerns, and the complexity of annotating subjective human behaviours that differ across cultures. Moreover, specialised evaluation metrics are needed to assess the effectiveness of these systems in capturing nuanced social trends. To advance multimodal AI in social science, efforts should focus on expanding data diversity, addressing privacy challenges, developing robust annotation and evaluation methods, and fostering robust ethical frameworks and cross-disciplinary expertise. These efforts will ensure that the insights generated are fair, transparent, and aligned with societal goals.

\subsection*{Engineering}
Beyond self-driving cars, multimodal AI supports advanced autonomous systems, including robotics, precision manufacturing, and medical applications. Robots equipped with multimodal AI can integrate vision, haptic feedback, and sensor data to perform delicate tasks, such as in surgical procedures \cite{he2022robotic} or complex manipulation \cite{wang2024multimodal}, where precision is paramount. Additionally, multimodal AI can enable seamless human-robot interaction by  integrating speech, vision, and environmental understanding \cite{duncan2024survey}. Speech technology and natural language processing can enhance user interfaces, improving interaction while minimising distractions for operators in robotic applications.

Despite the potential of multimodal AI, challenges remain in achieving effective data fusion, system interoperability, and robust model generalisation. Furthermore, the lack of diverse datasets, particularly for edge cases or atypical deployment environments, limits the scalability of these systems. Establishing open platforms for sharing data and toolkits can accelerate innovation and ensure interoperability across engineering applications. Integrating insights from legal studies and social sciences can enhance compliance with evolving regulations and improve user-centred design, boosting societal acceptance of autonomous systems. Environmental science can contribute to the development of energy-efficient and sustainable designs, particularly for robotics and manufacturing processes. Addressing these challenges requires interdisciplinary collaborations to improve data diversity and develop scalable, interoperable systems. 


\subsection*{Science}
Multimodal AI is transforming scientific discovery by integrating diverse data sources to model complex phenomena. Beyond its role in vehicle dynamics simulation, as seen in self-driving car design, multimodal AI is accelerating advancements in materials science. For instance, integrating experimental and computational data enables the development of improved batteries, fuel cells and supercapacitors \cite{liu2024recent}. We can incorporate domain knowledge via techniques such as physics-informed neural networks \cite{raissi2019physics,karniadakis2021physics} and neurosymbolic AI \cite{garcez2023neurosymbolic} to improve the accuracy of scientific simulations and enhance model trustworthiness and predictive power across applications \cite{yang2023context, rezaei2021learning}. 

Remaining challenges of multimodal AI for scientific discovery lie in integrating heterogeneous data, ranging from atomic-level properties to macroscopic observations, incorporating synthetic data, and addressing the scarcity of high-fidelity datasets amidst an abundance of low-fidelity data. Advancing scientific discovery with multimodal AI requires comprehensive datasets, rigorous data standards, and interdisciplinary research environments. These efforts will ensure that models are aligned with real-world dynamics, unlocking new possibilities in materials science, physics, and beyond \cite{trinh2024olympiad,raayoni2021ramanujan}.


\subsection*{Sustainability}
Multimodal AI can help address sustainability challenges, such as biodiversity conservation and environmental monitoring \cite{gui2024remote,van2025monitoring}. Beyond its applications in climate change adaptation, multimodal AI can combine geospatial data, bioacoustic recordings, and environmental DNA to track species and ecosystems \cite{pollock2025harnessing}, supporting conservation planning and enhancing understanding of ecosystem dynamics. In terrestrial environments, ecoacoustic and LiDAR data can be combined to model biodiversity variation across complex landscapes \cite{rappaport2020acoustic}. In marine settings, integrating remote sensing with in situ sensors enables long-term monitoring of ecosystem health and debris pathways \cite{maximenko2021integrated}.

Geographic imbalances in data availability, particularly in underrepresented regions, limit model fairness and accuracy. Marine environments pose unique sensing challenges \cite{briciu2023sensors}, including high turbidity and limited optical visibility, which limit the effectiveness of conventional imaging approaches. This necessitates innovative solutions such as autonomous underwater vehicles and acoustic sensors to access deeper waters, remote habitats, and ecologically sensitive regions \cite{whitt2020future}. Future advancements will depend on building diverse, scalable datasets, de-biasing models, and leveraging digital twins \cite{Li2023} for real-time monitoring and actionable insights. By addressing these challenges, multimodal AI can make meaningful contributions to global sustainability efforts \cite{zhao2024artificial}.


\subsection*{Finance}
Multimodal AI can advance finance by integrating financial, environmental, and social data to improve risk assessments, market forecasting, and sustainable investment strategies. In particular, it can enhance ESG (environmental, social, and governance) investing by helping stakeholders make more responsible financial decisions \cite{ang2024learning}. Integrating financial data and environmental data, such as satellite imagery, can enable a more comprehensive approach to risk management and resilience \cite{leng2024power}. Additionally, graph-based multimodal models can monitor systemic risks by capturing complex relationships within financial ecosystems \cite{ang2024learning}, supporting the early detection of market disruptions. Traditional financial models often struggle to respond to economic shocks such as natural disasters. Integrating real-time, multimodal data can improve model adaptability and accuracy under volatile conditions, ultimately enhancing financial stability \cite{cao2023ai}.



Challenges such as data privacy, regulatory complexity, and the need for explainability continue to be substantial barriers. Addressing these issues requires comprehensive multimodal datasets and transparent evaluation frameworks to build trust and ensure compliance with regulations. Interdisciplinary collaborations can help refine multimodal AI systems in the finance sector. Environmental science informs environmental risk assessments, enabling more resilient investment strategies and credit evaluations in the face of severe adverse events. Social sciences provide socio-political insights that promote transparency, equity, and ethical compliance, ensuring that financial practices align with societal goals. These collaborations align financial systems with ethical and sustainable practices while equipping stakeholders to make resilient decisions in an evolving financial landscape.

\section*{S2. Methods and extended analysis of the multimodal AI landscape}

\renewcommand{\figurename}{Supplementary Figure}
\setcounter{figure}{0}
\renewcommand{\thefigure}{\arabic{figure}}
\begin{figure*}[t]
\centering
    \includegraphics[width=\textwidth]{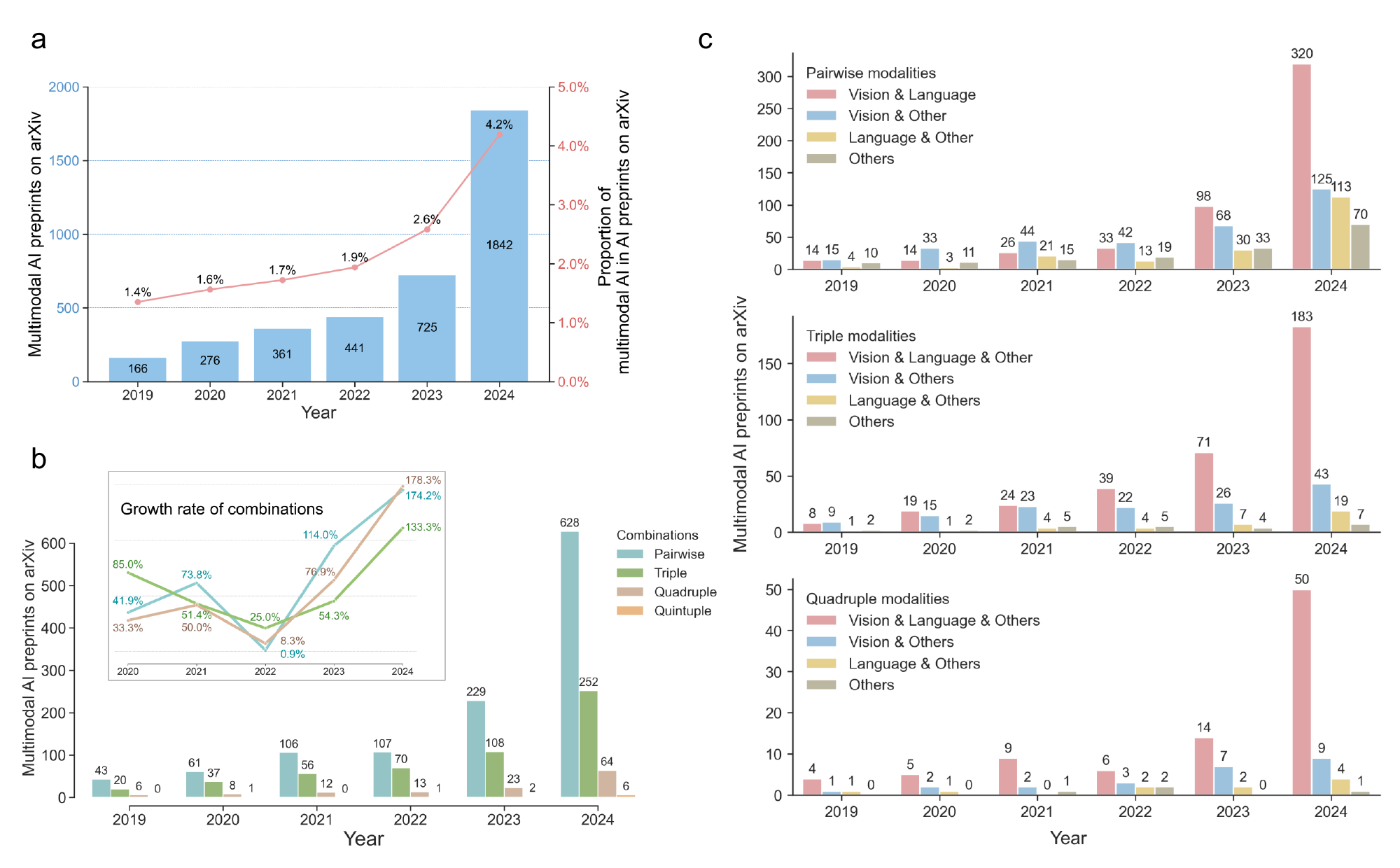}
    \caption{\textbf{Landscape of multimodal AI research  (2019--2024).} 
    {\color{myblue}
    \textbf{a,}~Yearly growth in the number of multimodal AI preprints on arXiv (bars) and their proportion among all AI preprints (line), both showing an accelerating upward trend since 2022, reflecting the field’s rapid expansion, likely driven by the large language model (LLM) revolution. \textbf{b,}~Distribution of multimodal AI preprints by the number of combined modalities. The inset line plot shows the growth rate of each combination. While pairwise combinations remain the most prevalent, combinations involving more modalities are steadily gaining attention, reflecting increasing interest in richer data fusion for tasks requiring diverse information sources. \textbf{c,}~Detailed breakdown of pairwise, triple, and quadruple modality combinations, highlighting trends across four modality clusters: vision and language, vision and other(s), language and other(s), and others. Combinations involving vision and language remain dominant. In particular, the number of pairwise combinations involving language has increased substantially from 2023 to 2024, likely due to the LLM-driven surge in multimodal research.}}   
\label{fig:modality_combinations}
\end{figure*}

All arXiv preprints \cite{ginsparg2011arxiv} from 2019 to 2024 were analysed to examine the latest trends in multimodal AI research. ArXiv was chosen for its open access, broad adoption within the AI community, and ability to reflect emerging developments more rapidly than peer-reviewed platforms. 
Raw data were extracted from the publicly available arXiv metadata dump \cite{arxiv_org_submitters_2024} hosted on Kaggle and filtered using queries for six general AI terms ``AI'', ``A.I.'', ``artificial intelligence'', ``machine learning'', ``deep learning'', and ``neural network'' appearing in either the title or the abstract to identify relevant preprints. The list was further refined to identify multimodal AI preprints by searching for ``multimodal'' and ``multi-modal'', and specific modalities were identified through targeted queries: ``vision'', ``image'', ``video'', and ``visual'' for vision; ``text'', ``language'', and ``textual'' for language; and similar sets of terms for the other six modalities: time series, graph, audio, spatial, sensor, and tabular data (see Data availability). As these queries are inherently approximate, conclusions are confined to overall trends supported by strong evidence rather than specific numerical details.

{\color{myblue}Supplementary Figure \ref{fig:modality_combinations} provides more detailed analyses of the multimodal AI landscape, including statistics on triple and quadruple modality combinations.}

\putbib
\end{bibunit}

\label{LastPageSupp}

\end{document}